\RequirePackage{amsmath}
\documentclass[runningheads]{llncs}
\usepackage{graphicx}
\usepackage{multirow}
\usepackage{bm}
\usepackage{amsfonts,amssymb}
\usepackage{url}
\urldef{\mailsa}\path|{chunweim, zhanghex, mgao8}@buffalo.edu|

\begin{document}
\title{Neural Style Transfer Improves 3D Cardiovascular MR Image Segmentation on Inconsistent Data}
\titlerunning{Neural Style Transfer Improves 3D Cardiovascular MR Image Segmentation} %

\author{
Chunwei Ma\and Zhanghexuan Ji\and Mingchen Gao}
\authorrunning{C. Ma et al.} 
\institute{
Department of Computer Science and Engineering,
University at Buffalo,\\
The State University of New York, Buffalo, USA\\
\mailsa\\
}

\maketitle
\begin{abstract}
Three-dimensional medical image segmentation is one of the most important problems in medical image analysis and plays a key role in downstream diagnosis and treatment. Recent years, deep neural networks have made groundbreaking success in medical image segmentation problem. However, due to the high variance in instrumental parameters, experimental protocols, and subject appearances, the generalization of deep learning models is often hindered by the inconsistency in medical images generated by different machines and hospitals. In this work, we present StyleSegor, an efficient and easy-to-use strategy to alleviate this inconsistency issue. Specifically, neural style transfer algorithm is applied to unlabeled data in order to minimize the differences in image properties including brightness, contrast, texture, etc. between the labeled and unlabeled data. We also apply probabilistic adjustment on the network output and integrate multiple predictions through ensemble learning. On a publicly available whole heart segmentation benchmarking dataset from MICCAI HVSMR 2016 challenge, we have demonstrated an elevated dice accuracy surpassing current state-of-the-art method and notably, an improvement of the total score by 29.91\%. StyleSegor is thus corroborated to be an accurate tool for 3D whole heart segmentation especially on highly inconsistent data, and is available at https://github.com/horsepurve/StyleSegor.

\end{abstract}

\section{Introduction}
The segmentation of 3D cardiac magnetic resonance (MR) images is the prerequisite for downstream diagnosis and treatment including heart disease identification and surgical planning. And there has been intensive research on the automatic algorithms for this segmentation problem, for purpose of alleviating the arduous manual labeling. Deep neural networks have made tremendous achievement on this task and many different architectures have been proposed, such as 3D U-Net~\cite{cciccek20163d}, VoxResNet~\cite{chen2018voxresnet}, 3D-DSN~\cite{dou20173d}, DenseVosNet~\cite{yu2017automatic}, VFN~\cite{xia18}, and their ensemble meta-learner~\cite{zheng2018new}, which improved the segmentation performance to the dice score of myocardium at $\sim$0.833 and that of blood pool at $\sim$0.939. 

\begin{figure}
\includegraphics[width=\textwidth]{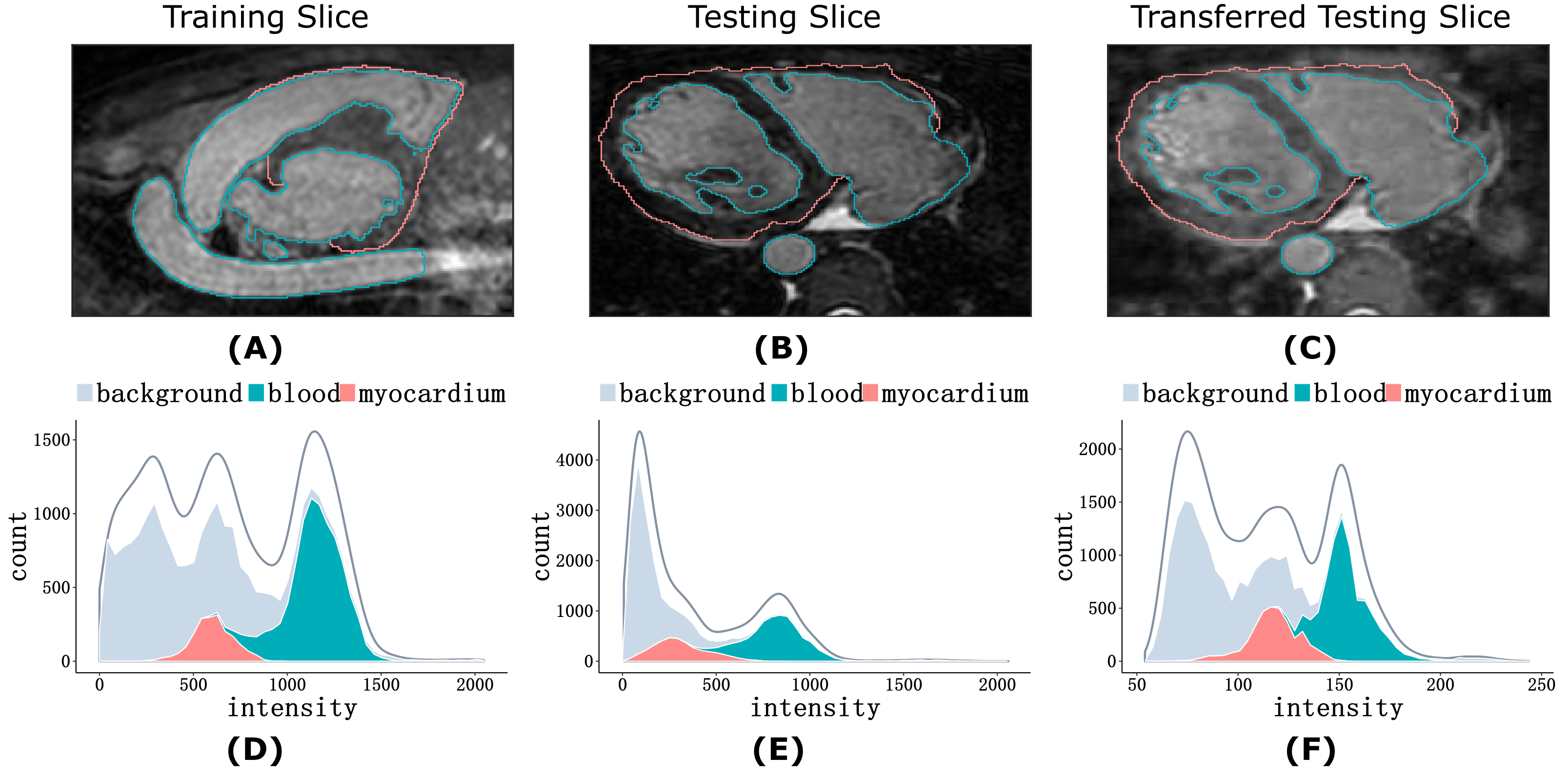}
\caption{Two representative slices from training (A) and testing set (B) and the transferred testing slice (C) are shown. Their intensity distributions are presented in D, E, and F, in which the intensity distributions of background, myocardium and blood pool are illustrated by gray, red, and green areas, respectively, and the gray lines show overall intensity distributions.} \label{fig3}
\end{figure}

However, the current accuracy of 3D cardiovascular MR image segmentation is still not well satisfactory for wider practice due to several issues. First, the morphological variation within the HVSMR data, originated from a variety of congenital heart defects, leads to difficulty in segmentation. Dong et al.~\cite{dong2018unsupervised} proposed an unsupervised domain adaptation network to enforce prediction masks to be similar across domains. However, the shapes of myocardium and blood pool are much more complex than lungs in 2D X-rays images. More important, we have observed non-negligible inter-subject variation within the training and testing images, including brightness, resolution, texture, and signal to noise ratio. In HVSMR data, the training samples are generally of high quality while the quality of the testing samples is relatively low. In a training image (Fig.~\ref{fig3}A), the intensity distribution (gray line in Fig.~\ref{fig3}D) exhibits three distinguishable peaks whereas the testing image (Fig.~\ref{fig3}, B and E) shows a substantial overlap of myocardium signal and background signal. This dataset shift phenomena~\cite{5376} significantly hampered the generalization of deep neural network models. Zhao et al.~\cite{zhao2019data} proposed using learned transforms to generate samples used in data augmentation aiming at one-shot segmentation. In our preliminary experiments, we found that augmenting the training set with images generated from the low-quality domain contributed little to the overall performance.

To address these challenges, we propose StyleSegor, a novel pipeline for 3D MR image segmentation of cardiac and vascular structures. StyleSegor has three main advantages. First, we adopted atrous convolution network with atrous spatial pyramid pooling module as an efficient way to retain as many details of feature maps as possible and achieve better segmentation on subtle structures. Second, we leverage neural style transfer to minimize the inter-subject variation. Every slice sample in the testing data is directly transferred to the same style of a target from the training set. Third, in order to fully utilize both the original and the transformed image data, an ensemble learning scheme is developed through voting of multiple predictions. On the HVSMR 2016 challenge dataset, StyleSegor has demonstrated superior performance compared with other methods, and notably, an improvement of the total score by 29.91\%, showing the effectiveness of our strategy.

\section{Methods}
\begin{figure}
\includegraphics[width=\textwidth]{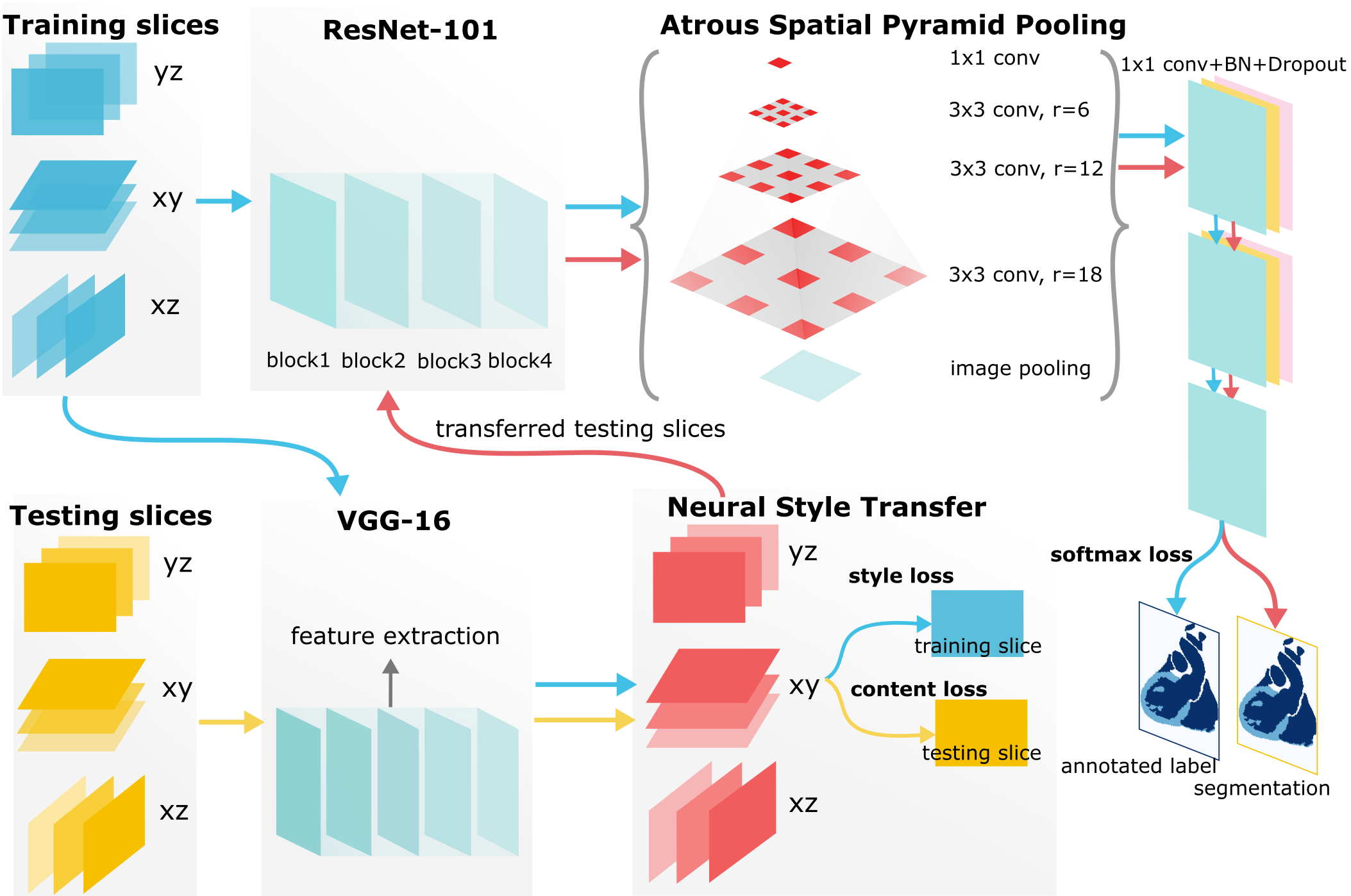}
\caption{Schematic illustration of StyleSegor workflow. A modified DeepLabv3 model with ResNet-101 backbone acts as our segmentation network and three models are trained in parallel for three planes $xy, yz, and\ zx$ (blue arrows). Meanwhile, testing slices are transferred to target styles guided by content loss and style loss, and then are fed into the segmentation network (red arrows).} \label{fig1}
\end{figure}

The complete pipeline of StyleSegor is shown in Fig.~\ref{fig1}. The standard ResNet-101 and VGG-16 networks serve as the backbones for segmentation and style transfer, respectively. The network is pre-trained on the combination of images from three orthogonal planes and then fine-tuned separately on images from each plane. Each testing slice goes through style transfer network to generate its transferred counterpart, which is in turn segmented using the fine-tuned segmentation model.

\subsection{Atrous Convolutional Neural Network for Dense Image Segmentation}
For our baseline model, we modified DeepLabv3~\cite{Chen_2018_ECCV}, the state-of-the-art 2D semantic segmentation network, with ResNet-101 backbone. In order to fully utilize multi-scale information of feature maps elicited from ResNet, a pyramid of atrous convolution layer with various atrous rates $r=(6,12,18)$ is constructed on top of the last block of ResNet. Besides the three atrous convolution layers, the features from a $1\times1$ convolution and a bilinearly upsampled duplication of the input feature map are also considered. These 5 layers compose the atrous spatial pyramid pooling (ASPP) module whose feature maps are all concatenated. Finally, three $1\times1$ convolution layers are used to generate the final logits. Two batch normalization layers and two dropout layers (dropout rates being 0.5 and 0.1) are inserted between the final 3 convolution layers.

\subsection{Neural Style Transfer on Inconsistent Data}
Due to the large inconsistency between the training and testing data (Fig.~\ref{fig2} A), we apply a neural style transfer algorithm~\cite{Gatys_2016_CVPR} on all 10 testing samples. Specifically, two types of loss, content loss and style loss are optimized, to change the style of a testing slice $\bm{x}$ to be similar to that of a target training slice $\bm{y}$, while simultaneously impose constraint on the generated image $\bm{\hat{y}}$ to maintain its content. Formally, given a feature extraction network $\phi$ that has $J$ layers generating feature maps, the content loss is written as
\begin{equation}
\ell_{content}^{\phi}(\bm{\hat{y}},\bm{x}) =  \sum_{j=1}^{J} \frac{1}{C_j H_j W_j} \parallel \phi_j(\bm{\hat{y}}) - \phi_j(\bm{x}) \parallel_2^2,
\end{equation}
\noindent where $C_j, H_j, W_j$ are the dimensions of the feature maps in the $j^{th}$ layer.

On the other hand, in order to measure the discrepancy between the generated slice $\bm{\hat{y}}$ and target slice $\bm{y}$, Gram matrix, originally designed to capture texture information, is to be computed. The $j^{th}$ Gram matrix for $\bm{y}$ is
\begin{equation}
G_j^{\phi}(\bm{y})_{i,k} = \frac{1}{C_j H_j W_j} \sum_{h=1}^{H_j} \sum_{w=1}^{W_j} \phi_j(\bm{y})_{h,w,i} \phi_j(\bm{y})_{h,w,k},
\end{equation}
\noindent that is, the $i,k$ position at the $j^{th}$ Gram matrix measures the correlation (inner product) of the $i^{th}$ and the $k^{th}$ feature maps in the $j^{th}$ layer. Subsequently, the style loss is
\begin{equation}
\ell_{style}^{\phi}(\bm{\hat{y}},\bm{y}) = \sum_{j=1}^{J}  \parallel G_j^{\phi}(\bm{\hat{y}}) - G_j^{\phi}(\bm{y}) \parallel_2^2.
\end{equation}
\noindent The total loss is a weighted combination of content loss ans style loss
\begin{equation}
\ell_{total}^{\phi}(\bm{\hat{y}},\bm{x},\bm{y}) = \alpha \ell_{content}^{\phi}(\bm{\hat{y}},\bm{x}) + \beta \ell_{style}^{\phi}(\bm{\hat{y}},\bm{y}),
\end{equation}
\noindent where $\alpha$ and $\beta$ are user-specified hyper parameters to adjust the relative weights of the two losses. During the style transfer process, stochastic gradient descent (SGD) optimization is directly applied on the generated image $\bm{\hat{y}}$ starting from the content image $\bm{x}$.

Until now, a remaining question is, for a given testing slice, how to find the optimal training slice as its target slice. We address this problem in several steps. First, the pairwise similarities of all training and testing samples are measured through the $1^{st}$ Wasserstein metric
\begin{equation} 
W(r,g) = \inf_{\gamma \in \Gamma(r,g)} \mathbb{E}_{(x,y) \sim \gamma} \parallel x-y \parallel,
\end{equation}
\noindent where $\Gamma(r,g)$ denotes the set of all joint distributions $\gamma(x,y)$ whose marginals are $r,g$, which measures the work needed to transport from x to y with optimal transport plan. Considering the high difference in intensity ranges across samples, Wasserstein distance is a suitable indicator for sample similarity. Based on these similarities, all samples are clustered using hierarchical clustering algorithm~\cite{2011modern}, and the training samples reside in one cluster serve as the style library (the first cluster in Fig.~\ref{fig2}A). Using our baseline network, the percentages of the three labels within each testing slice are used to measure the distance between two slices, and the slice in the style library with the smallest Euclidean distance to the testing slice is chosen as the target style. 

Because in StyleSegor, a full training process is required for every content-style pair, we use VGG-16, a lightweight network, as the feature extraction network $\phi$ and the feature maps after the $2^{nd}, 4^{th}, 7^{th}, 10^{th}$ convolution layers are used to compute the Gram matrices (see Fig.~\ref{fig1}).

\subsection{Probabilistic Adjustment and Ensemble Learning}
Based on the observation that the signals of myocardium and blood pool tend to be overwhelmed by the background signal (Fig.~\ref{fig2}, E and H), we perform a probabilistic adjustment step and adjust the score for one label at position $i$ by conditioning on the scores of other labels
\begin{equation}
c(p_k) = \mathop{\arg\max}_{k\in(1,2,3)} p_k \prod_{j \neq k} (1 - \frac{e^{p_j}}{\sum_{q\in(1,2,3)}{e^{p_q}}}),
\end{equation}
\noindent where $p_k, k \in (1,2,3)$ is the logits output from the network for three labels. For example, the score of myocardium is multiplied by the probabilities of both non-blood pool and non-background (Fig.~\ref{fig2}, F and I).

In machine learning practice, model ensemble is oft-used to take advantage of multiple models and predictions. Here we adopt a voting scheme to integrate segmentations obtained from both original and transformed images. The final label at position $i$ is the voting of $c(p_k^{(xy)}), c(p_k^{(yz)}), c(p_k^{(zx)})$ and $c(\sum_{xy, yz, zx} p_k)$ derived from the original images and $c(p_k^{'(xy)}), c(p_k^{'(yz)}), c(p_k^{'(zx)})$ and $c(\sum_{xy, yz, zx} p_k^{'})$ derived from the transferred images (Fig.~\ref{fig2}J).

\section{Experimental Results}
\subsubsection{Dataset and Training Process.}
\begin{figure}
\begin{center}
    \includegraphics[width=\textwidth]{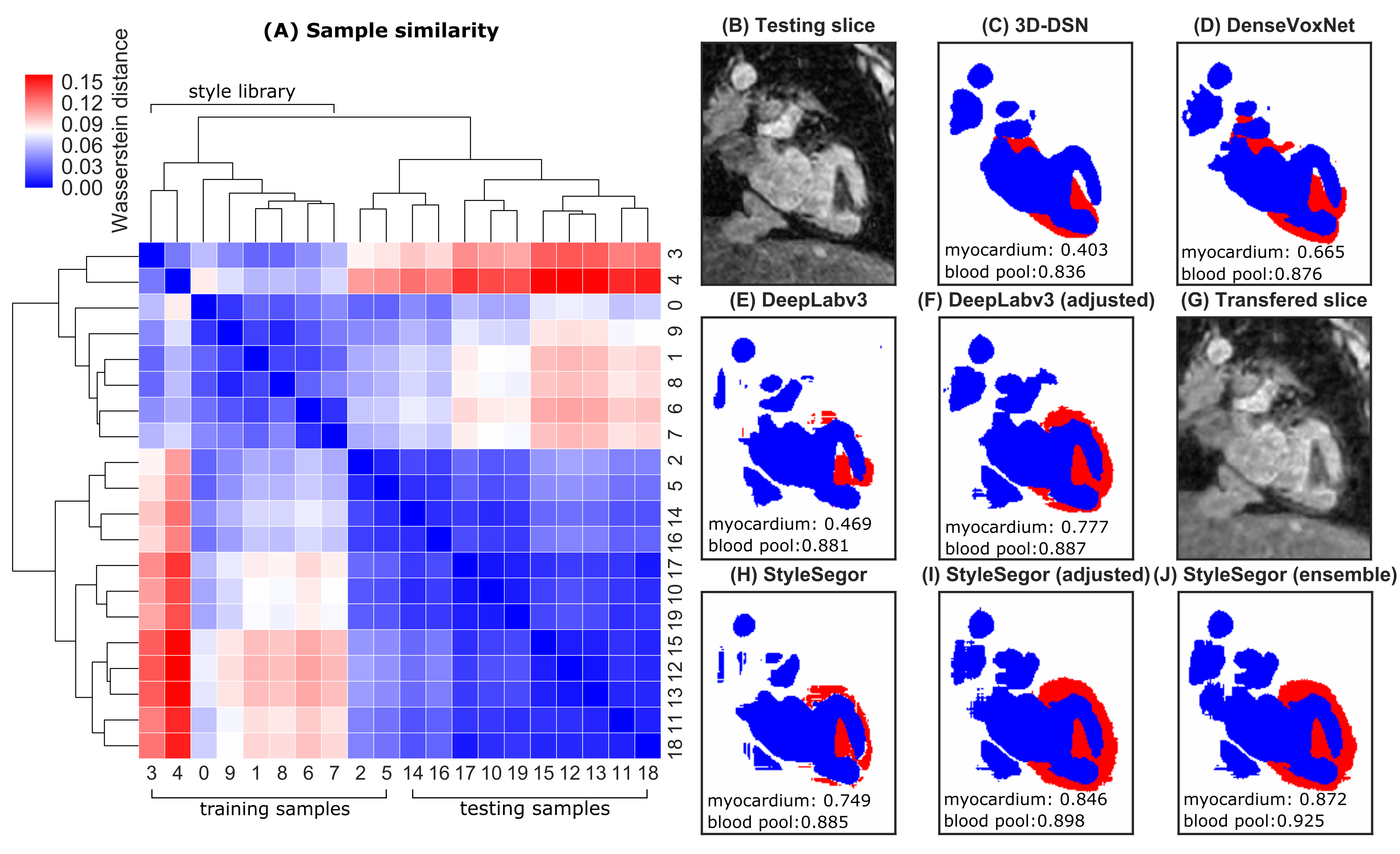}
    \caption{Clustering result of the 20 samples (A) and segmentation results on a representative testing slice (B to J). The blue and red colors in B to J represent blood pool and myocardium, respectively. The dice scores of the two labels are also shown. The ground truth labels for testing data are not made publicly available.} \label{fig2}
\end{center}
\end{figure}

We evaluate the performance of StyleSegor on HVSMR, the dataset for MICCAI 2016 Challenge on Whole-Heart and Great Vessel Segmentation from 3D Cardiovascular MRI in Congenital Heart Disease. Imaging was done in an axial view on a 1.5T scanner. Ten 3D MR scans, as well as the manually labeled annotations for myocardium and great vessel, are provided for training, but the labels for 10 testing scans are not made publicly available for fair comparison. After carefully investigating the properties of testing images, we observed that the signal of myocardium in testing samples is especially lower than in training samples (see Fig.~\ref{fig3}， A, D and B, E). The clustering result of training and testing samples based on Wasserstein metric is shown in Fig.~\ref{fig2}A, where training samples are marked from 0 to 9 and testing samples from 11 to 19. Clearly, all testing sample reside in the same cluster, which is significantly different from another cluster of training samples. In our style transfer network, the weights of style and content loss $\alpha$ and $\beta$ are set at $10^6$ and $1$, respectively, and the optimization terminates after 50 epochs, which typically takes 3s for one content-style pair on a GTX 1080 Ti card. The VGG-16 network is trained on ImageNet dataset.

To fully make use of the slices from three orthogonal planes, all slices are collected for training for 20 epochs with learning rate starting at 0.01. Then the slices derived from $xy$, $yz$, and $zx$ planes are used to fine-tune the model separately with learning rate starting from 0.002 for another 20 epochs each. A poly learning rate policy is employed where the starting learning rate is reduced by multiplying $(1-\frac{epoch}{max\_epoch})$. To accelerate the training process, the segmentation network is pre-trained on COCO dataset. During the training of our baseline model, a series of data augmentation strategy is applied. Each original image is randomly scaled with the rates ranged from 0.5 to 2.0 and a $480\times480$ patch is cropped then goes through random left-right flipping and random Gaussian blurring. Because the training images are randomly scaled during training, in testing process, each testing image is scaled with scaling rate $=(0.5, 0.75, 1, 1.25, 1.5, 2.0)$ and the accumulated score map is used to produce final segmentation.

A representative testing slice and the transferred slice of it are shown in Fig.~\ref{fig3}, B and C, while the intensity distribution of background, blood pool, and myocardium are illustrated in E and F. Interestingly, after style transfer, not only the brightness, contrast, texture of the image but also the distribution of the three labels are transformed to be very similar to the training image, and the myocardium signal is smartly elevated. 
\begin{table}[ht]
\centering
\caption{Comparison of different methods on HVSMR 2016 dataset. The weights of relative contributions of Dice, Average distance boundary (ADB), and Hausdorff distance to the Overall score are 0.5, -0.25, and -0.03, respectively.}
\label{tab1}
    \resizebox{\textwidth}{!}{
    \begin{tabular}{c|ccc|ccc|c}
    \hline
    \multirow{2}{*}{Method} & \multicolumn{3}{c|}{Myocardium} & \multicolumn{3}{c|}{Blood pool} & \multirow{2}{*}{\begin{tabular}[c]{@{}c@{}}Overall\\ score\end{tabular}} \\ \cline{2-7}
     & Dice & ADB {[}mm{]} & Hausdorff {[}mm{]} & Dice & ADB {[}mm{]} & Hausdorff {[}mm{]} &  \\ \hline
    3D U-Net~\cite{cciccek20163d} & 0.694$\pm$0.076 & 1.461$\pm$0.397 & 10.221$\pm$4.339 & 0.926$\pm$0.016 & 0.940$\pm$0.192 & 8.628$\pm$3.390 & -0.419 \\
    3D DSN~\cite{dou20173d} & 0.739$\pm$0.072 & 1.035$\pm$0.240 & 5.248$\pm$1.332 & 0.928$\pm$0.014 & 1.017$\pm$0.181 & 7.704$\pm$2.892 & -0.162 \\
    VoxResNet~\cite{chen2018voxresnet} & 0.774$\pm$0.067 & 1.026$\pm$0.400 & 6.572$\pm$0.013 & 0.929$\pm$0.013 & 0.981$\pm$0.186 & 9.966$\pm$3.021 & -0.202 \\
    DenseVoxNet~\cite{yu2017automatic} & 0.821$\pm$0.041 & 0.964$\pm$0.292 & 7.294$\pm$3.340 & 0.931$\pm$0.011 & 0.938$\pm$0.224 & 9.533$\pm$4.194 & -0.161 \\
    Wolterink et. al~\cite{Wolterink} & 0.802$\pm$0.060 & 0.957$\pm$0.302 & 6.126$\pm$3.565 & 0.926$\pm$0.018 & 0.885$\pm$0.223 & 7.069$\pm$2.857 & -0.036 \\
    VFN~\cite{xia18} & 0.773$\pm$0.098 & 0.877$\pm$0.318 & 4.626$\pm$2.319 & 0.935$\pm$0.009 & 0.770$\pm$0.098 & 5.420$\pm$2.152 & 0.108 \\
    Zheng et. al~\cite{zheng2018new} & 0.833$\pm$0.054 & {\bfseries0.681$\pm$0.178} & 3.285$\pm$1.370 & {\bfseries0.939$\pm$0.008} & 0.733$\pm$0.143 & 5.670$\pm$2.808 & 0.234 \\ \hline
    DeepLabv3 (baseline) & 0.648$\pm$0.156 & 1.234$\pm$0.531 & 5.960$\pm$3.921 & 0.920$\pm$0.025 & 0.983$\pm$0.309 & 7.343$\pm$2.999 & -0.214 \\
    StyleSegor (baseline) & 0.744$\pm$0.085 & 1.061$\pm$0.322 & 5.610$\pm$2.641 & 0.923$\pm$0.022 & 1.000$\pm$0.285 & 5.778$\pm$2.999 & -0.061 \\
    DeepLabv3 (adjusted) & 0.808$\pm$0.057 & 0.820$\pm$0.230 & 3.105$\pm$1.033 & 0.916$\pm$0.018 & 1.038$\pm$0.227 & 7.887$\pm$2.787 & 0.031 \\
    StyleSegor (adjusted) & 0.825$\pm$0.031 & 0.934$\pm$0.237 & 4.633$\pm$2.241 & 0.923$\pm$0.014 & 1.073$\pm$0.191 & 7.435$\pm$2.649 & -0.030 \\
    StyleSegor (ensemble) & {\bfseries0.839$\pm$0.037} & 0.689$\pm$0.140 & {\bfseries2.832$\pm$0.660} & 0.937$\pm$0.014 & {\bfseries0.731$\pm$0.182} & {\bfseries4.023$\pm$1.299} & {\bfseries0.304} \\ \hline
    \end{tabular}}
\end{table}

\subsubsection{Quantitative Comparisons.}
The comparison of StyleSegor and our baseline network, along with other segmentation methods are shown in Table~\ref{tab1}, and the visualization of those segmentation results is provided in Fig.~\ref{fig2}, B to J. After probabilistic adjustment, although our baseline model only performs 2D convolution, it comes up with satisfactory segmentation with dice score of myocardium at 0.808 and that of blood pool at 0.919, and notably, by virtue of the large field of view, it produces the best Hausdorff distance at 3.105 mm compared with previous methods. After style transfer, the segmentation performance of myocardium is promoted to 0.825 and that of blood pool to 0.923, suggesting that with the promotion of myocardium signal, the myocardium structures are better recognized by the same model. However, we notice that after transfer, the Hausdorff distance of myocardium segmentation is enlarged to 4.633 mm, probably caused by false positive prediction of myocardium label brought by style transfer. And this false positive prediction is likely to be eliminated by the ensemble of multiple predictions. As shown in the last row of Table~\ref{tab1}, the ensemble result is better than either StyleSegor or DeepLabv3, with dice score of myocardium at 0.839 and that of blood pool at 0.937. Notably, the Hausdorff distances are greatly minimized to 2.832 mm for myocardium and 4.023 mm for blood pool, and the overall score is boosted to 0.304, a 29.91\% improvement compared with previously best result, demonstrating StyleSegor's strength to locate the region of interest.

\section{Conclusion}
In this paper, we present StyleSegor, a novel pipeline for 3D cardiac MR image segmentation. The neural style transfer algorithm automatically transfers the testing images towards the domain of training images, making them easier to be processed by the same model. Our StyleSegor pipeline is also easy to be used in other tasks such as disease detection and classification when data inconsistency is an inevitable issue, e.g., tasks involving datasets collected from different hospitals or institutions.

%
%
\bibliographystyle{splncs04}
\bibliography{refs}

\end{document}